\documentclass[11pt]{article}
\usepackage{eacl2017}
\usepackage{times}
\usepackage{url}
\usepackage{latexsym}

\usepackage{amsmath}
\usepackage{amsfonts}
\usepackage{mathtools}
\usepackage{bm}

\usepackage{booktabs,array}
\newcolumntype{C}[1]{>{\centering\arraybackslash}p{#1}}
\newcolumntype{L}[1]{>{\raggedright\arraybackslash}p{#1}}
\usepackage{multirow}

\usepackage[pdftex]{graphicx}
\graphicspath{{figures/}}

\usepackage{enumitem}
\setlist{parsep=0pt,listparindent=\parindent}

\allowdisplaybreaks

\eaclfinalcopy 

\title{Unsupervised Training for Large Vocabulary Translation\\Using Sparse Lexicon and Word Classes}

\author{Yunsu Kim,\; Julian Schamper \and Hermann Ney\\
  Human Language Technology and Pattern Recognition Group\\
  RWTH Aachen University\\
  {\tt \{surname\}@cs.rwth-aachen.de}}

\date{}

\begin{document}
\maketitle
\begin{abstract}
We address for the first time unsupervised training for a translation task with hundreds of thousands of vocabulary words. We scale up the expectation-maximization (EM) algorithm to learn a large translation table without any parallel text or seed lexicon. First, we solve the memory bottleneck and enforce the sparsity with a simple thresholding scheme for the lexicon. Second, we initialize the lexicon training with word classes, which efficiently boosts the performance. Our methods produced promising results on two large-scale unsupervised translation tasks.\vspace{1em}
\end{abstract}

\section{Introduction}
Statistical machine translation (SMT) heavily relies on parallel text to train translation models with supervised learning. Unfortunately, parallel training data is scarce for most language pairs, where an alternative learning formalism is highly in need.

In contrast, there is a virtually unlimited amount of monolingual data available for most languages. Based on this fact, we define a basic \emph{unsupervised learning problem for SMT} as follows; given only a source text of arbitrary length and a target side LM, which is built from a huge target monolingual corpus, we are to learn translation probabilities of all possible source-target word pairs.

We solve this problem using the EM algorithm, updating the translation hypothesis of the source text over the iterations. In a very large vocabulary setup, the algorithm has two fundamental problems: 1) A full lexicon table is too large to keep in memory during the training. 2) A search space for hypotheses grows exponentially with the vocabulary size, where both memory and time requirements for the forward-backward step explode.

For this condition, it is unclear how the lexicon can be efficiently represented and whether the training procedure will work and converge properly. This paper answers these questions by 1) filtering out unlikely lexicon entries according to the training progress and 2) using word classes to learn a stable starting point for the training. For the first time, we eventually enabled the EM algorithm to translate 100k-vocabulary text in an unsupervised way, achieving 54.2\% accuracy on \textsc{Europarl} Spanish$\rightarrow$English task and 32.2\% on \textsc{IWSLT} 2014 Romanian$\rightarrow$English task.

\section{Related Work}

Early work on unsupervised sequence learning was mainly for \emph{deterministic decipherment}, a combinatorial problem of matching input-output symbols with 1:1 or homophonic assumption \cite{Knight06,Ravi11a,Nuhn13}. \emph{Probabilistic decipherment} relaxes this assumption to allow many-to-many mapping, while the vocabulary is usually limited to a few thousand types \cite{Nuhn12,Dou13,Nuhn14,Dou15}.

There has been several attempts to improve the scalability of decipherment methods, which are however not applicable to 100k-vocabulary translation scenarios. For EM-based decipherment, \newcite{Nuhn12} and \newcite{Nuhn14} accelerate hypothesis expansions but do not explicitly solve the memory issue for a large lexicon table. Count-based Bayesian inference \cite{Dou12,Dou13,Dou15} loses all context information beyond bigrams for the sake of efficiency; it is therefore particularly effective in contextless deterministic ciphers or in inducing an auxiliary lexicon for supervised SMT. \newcite{Ravi13} uses binary hashing to quicken the Bayesian sampling procedure, which yet shows poor performance in large-scale experiments.

Our problem is also related to \emph{unsupervised tagging} with hidden Markov model (HMM). To the best of our knowledge, there is no published work on HMM training for a 100k-size discrete space. HMM taggers are often integrated with sparse priors \cite{Goldwater07,Johnson07}, which is not readily possible in a large vocabulary setting due to the memory bottleneck.

Learning a good initialization on a smaller model is inspired by \newcite{Och03} and \newcite{Knight06}. Word classes have been widely used in SMT literature as factors in translation \cite{Koehn07,Rishoj11} or smoothing space of model components \cite{Wuebker13,Kim16}.

\section{Baseline Framework}
Unsupervised learning is yet computationally demanding to solve general translation tasks including reordering or phrase translation. Instead, we take a simpler task which assumes 1:1 monotone alignment between source and target words. This is a good initial test bed for unsupervised translation, where we remove the reordering problem and focus on the lexicon training.

Here is how we set up our unsupervised task: We rearranged the source words of a parallel corpus to be monotonically aligned to the target words and removed multi-aligned or unaligned words, according to the learned word alignments. The corpus was then divided into two parts, using the source text of the first part as an input ($f_1^N$) and the target text of the second part as LM training data. In the end, we are given only monolingual part of each side which is not sentence-aligned. The statistics of the preprocessed corpora for our experiments are given in Table \ref{tab:corpus-stat}.

\begin{table}[!ht]
  \centering
  \begin{tabular}{cccc}
    \toprule
     & & Source & Target\\
    Task & & (Input) & (LM)\\
    \midrule
    \textsc{EuTrans} & \multicolumn{1}{|c}{Run. Words} & 85k & 4.2M\\
    es-en & \multicolumn{1}{|c}{Vocab.} & 677 & 505\\
    \midrule
    \textsc{Europarl} & \multicolumn{1}{|c}{Run. Words} & 2.7M & 42.9M\\
    es-en & \multicolumn{1}{|c}{Vocab.} & 32k & {\bf 96k}\\
    \midrule
    \textsc{IWSLT} & \multicolumn{1}{|c}{Run. Words} & 2.8M & 13.7M\\
    ro-en & \multicolumn{1}{|c}{Vocab.} & {\bf 99k} & {\bf 114k}\\
    \bottomrule
  \end{tabular}
\caption{Corpus statistics.}
\label{tab:corpus-stat}
\end{table}

To evaluate a translation output $\hat{e}_1^N$, we use token-level accuracy (Acc.):
\begin{align}
  \textrm{Acc.} = \frac{\sum\limits_{n=1}^N[\hat{e}_n=r_n]}{N}
  \label{eq:acc}
\end{align}
where $r_1^N$ is the reference output which is the target text of the first division of the corpus. It aggregates all true/false decisions on each word position, comparing the hypothesis with the reference. This can be regarded as the inverse of word error rate (WER) without insertions and deletions. It is simple to understand and nicely fits to our reordering-free task.

In the following, we describe a baseline method to solve this task. For more details, we refer the reader to \newcite{Schamper15}.

\subsection{Model}
We adopt a noisy-channel approach to define a joint probability of $f_1^N$ and $e_1^N$ as follows:
\begin{align}
  p(e_1^N,f_1^N) &= \prod_{n=1}^N p(e_n|e_{n-m+1}^{n-1})\,p(f_n|e_n) \label{eq:model}
\end{align}
which is composed of a pre-trained $m$-gram target LM and a word-to-word translation model. The translation model is parametrized by a full table over the entire source and target vocabularies:
\begin{align}
  p(f|e)=\theta_{f|e} \label{eq:lex-dense}
\end{align}
with normalization constraints $\forall_e \:\sum_f\theta_{f|e}=1$. Having this model, the best hypothesis $\hat{e}_1^N$ is obtained by the Viterbi decoding.

\subsection{Training}
To learn the lexicon parameters $\{\theta\}$, we use maximum likelihood estimation. Since a reference translation is not given, we treat $e_1^N$ as a latent variable and use the EM algorithm \cite{Dempster77} to train the lexicon model. The update equation for each maximization step (M-step) of the algorithm is:
\begin{align}
  \hat{\theta}_{f|e}=\frac{\sum\limits_{n:\,f_n=f}p_n(e|f_1^N)}{\sum\limits_{f'}\sum\limits_{n':\,f_{n'}=f'}p_{n'}(e|f_1^N)} \label{eq:lex-estimate}
\end{align}
with $p_n(e|f_1^N) = \sum_{e_1^N:e_n=e}\, p(e_1^N|f_1^N)$. This quantity is computed by the forward-backward algorithm in the expectation step (E-step).

\section{Sparse Lexicon}

Loading a full table lexicon (Equation \ref{eq:lex-dense}) is infeasible for very large vocabularies. As only a few $f$'s may be eligible translations of a target word $e$, we propose a new lexicon model which keeps only those entries with a probability of at least $\tau$:
\begin{align}
  \mathcal{F}(e)&=\{f\:|\:\hat{\theta}_{f|e}\ge\tau\}\\[-2.3em]
  \nonumber
\end{align}
\begin{align}
  p_\text{sp}(f|e)&=
  \begin{dcases}
    \frac{\hat{\theta}_{f|e}}{\sum\limits_{f'\in\mathcal{F}(e)}\hat{\theta}_{f'|e}} & \text{if }f\in\mathcal{F}(e)\\
    0 & \text{otherwise}
  \end{dcases} \label{eq:lex-sparse}
\end{align}
We call this model \emph{sparse} lexicon, because only a small percentage of full lexicon is \emph{active}, i.e. has nonzero probability.

The thresholding by $\tau$ allows flexibility in the number of active entries over different target words. If $e$ has little translation ambiguity, i.e. probability mass of $\theta_{f|e}$ is concentrated at only a few $f$'s, $p_\text{sp}(f|e)$ occupies smaller memory than other more ambiguous target words. For each M-step update, it reduces its size on the fly as we learn sparser E-step posteriors.

However, the sparse lexicon might exclude potentially important entries in early training iterations, when the posterior estimation is still not reliable. Once an entry has zero probability, it can never be recovered by the EM algorithm afterwards. A naive workaround is to adjust the threshold during the training, but it does not actually help for the performance in our internal experiments.

To give a chance to zero-probability translations throughout the training, we smooth the sparse lexicon with a backoff model $p_\text{bo}(f)$:
\begin{align}
  p(f|e)=\lambda\cdot p_\text{sp}(f|e) + (1-\lambda)\cdot p_\text{bo}(f) \label{eq:lex-combine}
\end{align}
where $\lambda$ is the interpolation parameter. As a backoff model, we use uniform distribution, unigram of source words, or Kneser-Ney lower order model \cite{Kneser95,Foster06}.

\setlength{\tabcolsep}{3.7pt}
\begin{table}[!h]
 \centering
 \hspace*{-4.4pt}
  \begin{tabular}{ccccc}
    \toprule
    & & & Acc. & Active\hspace{0.65cm}\\
    Lexicon & $\tau$ & $p_\text{bo}$ & [\%] & Entries [\%]\\
    \midrule
    Full & - & - & 70.2 & 100\\
    \midrule
    \multirow{7}{*}{Sparse} & 0.01\enspace\enspace & \multirow{5}{*}{Uniform} & 64.0 & 1.1\\
    & 0.005\enspace & & 69.0 & 2.7\\
    & 0.002\enspace & & {\bf 72.3} & {\bf 5.1}\\
    & 0.001\enspace & & 71.8 & 6.3\\
    & 0.0001 & & 70.1 & 9.1\\
    \cmidrule{2-5}
    & \multirow{2}{*}{0.002\enspace} & Unigram & 71.2 & 5.1\\
    & & Kneser-Ney & 72.1 & 5.1\\
    \bottomrule
  \end{tabular}
\caption{Sparse lexicon with different threshold values and backoff models ($\lambda=0.99$). Initialized with uniform distributions and trained for 50 iterations with a bigram LM. No pruning is applied.}
\label{tab:threshold-smoothing}
\vspace{-0.5em}
\end{table}

In Table \ref{tab:threshold-smoothing}, we illustrate the effect of the sparse lexicon with \textsc{EuTrans} Spanish$\rightarrow$English task \cite{Amengual96}, comparing to the existing EM decipherment approach (full lexicon). By setting the threshold small enough ($\tau=0.001$), the sparse lexicon surpasses the performance of the full lexicon, while the number of active entries, for which memory is actually allocated, is greatly reduced. For the backoff, the uniform model shows the best performance, which requires no additional memory. The time complexity is not increased by using the new lexicon.

\begin{figure}[!h]
  \centering
  \includegraphics[width=0.98\linewidth]{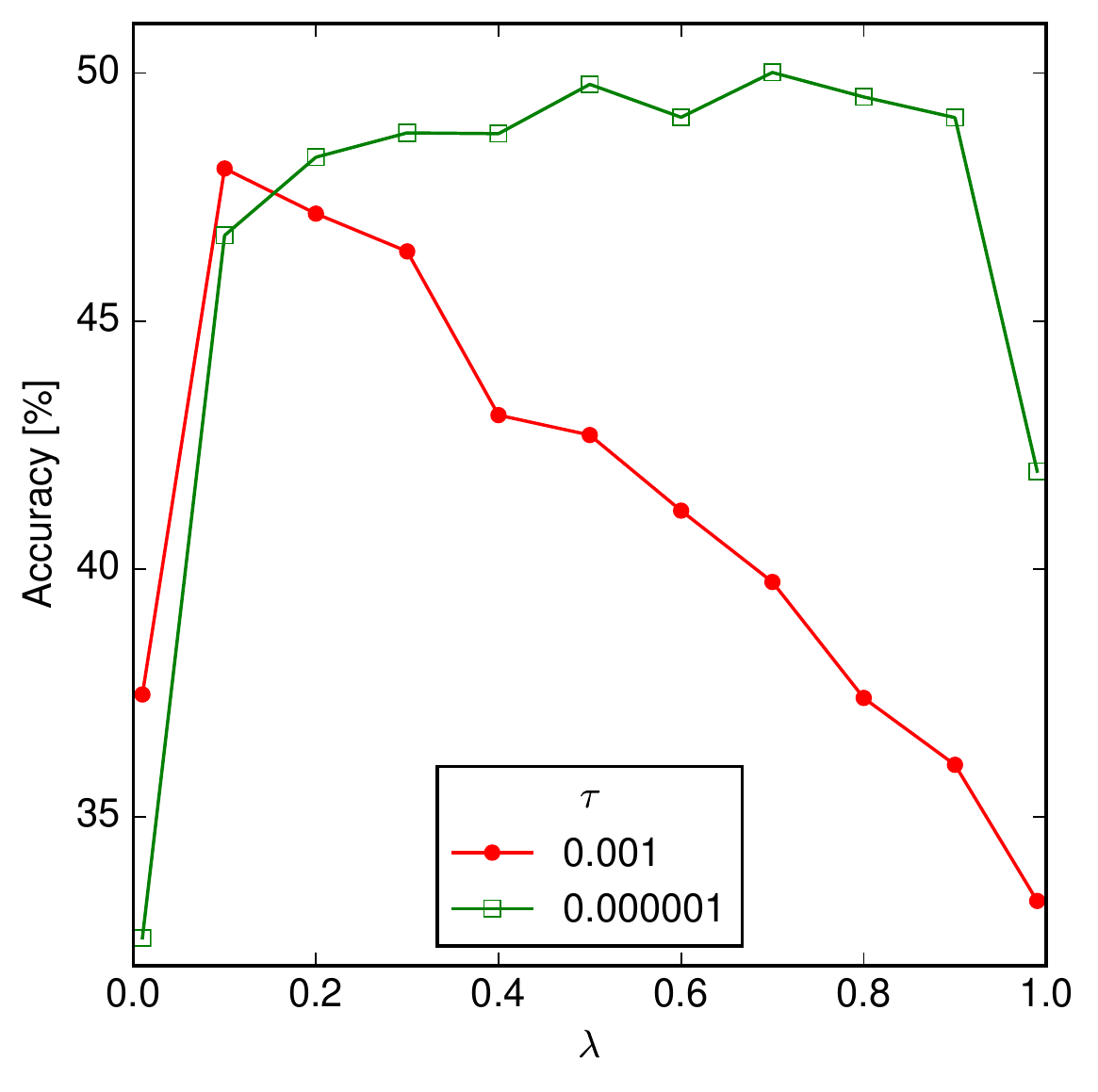}\vspace{-1.3em}
  \caption{Relation between sparse lexicon parameters (\textsc{Europarl} Spanish$\rightarrow$English task).}
  \label{fig:param}
  \vspace{-0.2em}
\end{figure}

We also study the mutual effect of $\tau$ and $\lambda$ (Figure \ref{fig:param}). For a larger $\tau$ (circles), where many entries are cut out from the lexicon, the best-performing $\lambda$ gets smaller ($\lambda=0.1$). In contrast, when we lower the threshold enough (squares), the performance is more robust to the change of $\lambda$, while a higher weight on the trained lexicon ($\lambda = 0.7$) works best. This means that, the higher the threshold is set, the more information we lose and the backoff model plays a bigger role, and vice versa.

The idea of filtering and smoothing parameters in the EM training is relevant to \newcite{Deligne95} and \newcite{Marcu02}. They leave out a fixed set of parameters for the whole training process, while we update trainable parameters for every iteration. \newcite{Nuhn14} also perform an analogous smoothing but without filtering, only to moderate the lattice pruning. Note that our work is distinct from the conventional pruning of translation tables in supervised SMT which is applied after the entire training.

\section{Initialization Using Word Classes}

Apart from the memory problem, it is inevitable to apply pruning in the forward-backward algorithm for runtime efficiency. The pruning in early iterations, however, may drop chances to find a better optimum in later stage of training. One might suggest to prune only for later iterations, but for large vocabularies, a single non-pruned E-step can blow up the total training time.

We rather stabilize the training by a proper initialization of the parameters, so that the training is less worsened by early pruning. We learn an initial lexicon on automatically clustered word classes \cite{Martin98}, following these steps:
\begin{enumerate}
\itemsep0.2em
\item Estimate word-class mappings on both sides ($\mathcal{C}_\text{src},\mathcal{C}_\text{tgt}$)
\item Replace each word in the corpus with its class
  \begin{align*}
    f &\mapsto \mathcal{C}_\text{src}(f)\\
    e &\mapsto \mathcal{C}_\text{tgt}(e)
  \end{align*}
\item Train a class-to-class full lexicon with a target class LM
\item Convert 3 to an unnormalized word lexicon by mapping each class back to its member words
  \begin{align*}
    \forall(f,e) \;\;\: q(f|e) := p(\mathcal{C}_\text{src}(f)|\,\mathcal{C}_\text{tgt}(e))
  \end{align*}
\item Apply the thresholding on 4 and renormalize (Equation \ref{eq:lex-sparse})
\end{enumerate}
where all $f$'s in an implausible source class are left out together from the lexicon. The resulting distribution $p_\text{sp}(f|e)$ is identical for all $e$'s in the same target class.

Word classes group words by syntactic or semantic similarity \cite{Brown92}, which serve as a reasonable approximation of the original word vocabulary. They are especially suitable for large vocabulary data, because one can arbitrarily choose the number of classes to be very small; learning a class lexicon can thus be much more efficient than learning a word lexicon.

\begin{table}[!h]
 \centering
  \begin{tabular}{cccc}
    \toprule
    \multicolumn{3}{c}{Initialization} & Acc. [\%]\\
    \midrule
    \multicolumn{3}{c}{Uniform} & 63.7\\
    \midrule
    & \#Classes & Class LM\\
    \midrule
    \multirow{5}{*}{\parbox{1.2cm}{\centering Word Classes}} & \multicolumn{1}{|c}{25} & 2-gram & 67.4 \\
    & \multicolumn{1}{|c}{50} & 2-gram & 69.1 \\
    & \multicolumn{1}{|c}{100} & 2-gram & 72.1 \\
    & \multicolumn{1}{|c}{50} & 3-gram & 76.0 \\
    & \multicolumn{1}{|c}{50} & 4-gram & {\bf 76.2} \\
    \bottomrule
  \end{tabular}
\caption{Sparse lexicon with word class initialization ($\tau=0.001$, $\lambda=0.99$, uniform backoff). Pruning is applied with histogram size 10.}
\label{tab:initialization}
\vspace{-0.5em}
\end{table}

Table \ref{tab:initialization} shows that translation quality is consistently enhanced by the word class initialization, which compensates the performance loss caused by harsh pruning. With a larger number of classes, we have a more precise pre-estimate of the sparse lexicon and thus have more performance gain. Due to the small vocabulary size, we are comfortable to use higher order class LM, which yields even better accuracy, outperforming the non-pruned results of Table \ref{tab:threshold-smoothing}. The memory and time requirements are only marginally affected by the class lexicon training.

Empirically, we find that the word classes do not really distinguish different conjugations of verbs or nouns. Even if we increase the number of classes, they tend to subdivide the vocabulary more based on semantics, keeping morphological variations of a word in the same class. From this fact, we argue that the word class initialization can be generally useful for language pairs with different roots. We also emphasize that word classes are estimated without any model training or language-specific annotations. This is a clear advantage for unknown/historic languages, where the unsupervised translation is indeed in need.

\section{Large Vocabulary Experiments}
\label{sec:lvexp}
We applied two proposed techniques to \textsc{Europarl} Spanish$\rightarrow$English corpus \cite{Koehn05} and \textsc{IWSLT} 2014 Romanian$\rightarrow$English TED talk corpus \cite{Cettolo12}. In the \textsc{Europarl} data, we left out long sentences with more than 25 words and sentences with singletons. For the \textsc{IWSLT} data, we extended the LM training part with news commentary corpus from \textsc{WMT} 2016 shared tasks.

We learned the initial lexicons on 100 classes for both sides, using 4-gram class LMs with 50 EM iterations. The sparse lexicons were trained with trigram LMs for 100 iterations ($\tau = 10^{-6}$, $\lambda = 0.15$). For further speedup, we applied per-position pruning with histogram size 50 and the preselection method of \newcite{Nuhn14} with lexical beam size 5 and LM beam size 50. All our experiments were carried out with the \textsc{Unravel} toolkit \cite{Nuhn15}.

Table \ref{tab:large-vocab} summarizes the results. The supervised learning scores were obtained by decoding with an optimal lexicon estimated from the input text and its reference. Our methods achieve significantly high accuracy with only less than 0.1\% of memory for the full lexicon. Note that using conventional decipherment methods is impossible to conduct these scales of experiments.

\setlength{\tabcolsep}{3.2pt}
\begin{table}[!ht]
  \centering
  \hspace*{-0.3em}
  \begin{tabular}{cccc}
    \toprule
     & \multicolumn{2}{c}{Acc. [\%]}\\
     \cmidrule(l{2pt}r{2pt}){2-3}
    Task & Supervised & Unsupervised & Lex. Size [\%]\\
    \midrule
    es-en & 77.5 & {\bf 54.2} & {\bf 0.06}\\
 	  ro-en & 72.3 & {\bf 32.2} & {\bf 0.03}\\
	\bottomrule
  \end{tabular}
\caption{Large vocabulary translation results.}
\label{tab:large-vocab}
\vspace{-0.5em}
\end{table}

\section{Conclusion and Future Work}

This paper has shown the first promising results on 100k-vocabulary translation with no bilingual data. To facilitate this, we proposed the sparse lexicon, which effectively emphasizes the multinomial sparsity and minimizes its memory usage throughout the training. In addition, we described how to learn an initial lexicon on word class vocabulary for a robust training. Note that one can optimize the performance to a given computing environment by tuning the lexicon threshold, the number of classes, and the class LM order.

Nonetheless, we still observe a substantial difference in performance between supervised and unsupervised learning for large vocabulary translation. We will exploit more powerful LMs and more input text to see if this gap can be closed. This may require a strong approximation with respect to numerous LM states along with an online algorithm.

As a long term goal, we plan to relax constraints on word alignments to make our framework usable for more realistic translation scenarios. The first step would be modeling local reorderings such as insertions, deletions, and/or local swaps \cite{Ravi11b,Nuhn12}. Note that the idea of thresholding in the sparse lexicon is also applicable to any normalized model components. When the reordering model is lexicalized, the word class initialization may also be helpful for a stable training.

\section*{Acknowledgments}

This work was supported by the Nuance Foundation and also received funding from the European Union's Horizon 2020 research and innovation programme under grant agreement n\textsuperscript{o}~645452 (QT21).

\bibliography{references}
\bibliographystyle{eacl2017}

\end{document}